\RequirePackage{silence}
\documentclass[pdflatex,sn-vancouver-num,iicol]{sn-jnl}

\usepackage[utf8]{inputenc}
\usepackage[T1]{fontenc}
\usepackage{graphicx}%
\usepackage{multirow}%
\usepackage{amsmath,amssymb,amsfonts}%
\usepackage{amsthm}%
\usepackage{mathrsfs}%
\usepackage[title]{appendix}%
\usepackage[dvipsnames]{xcolor}%
\usepackage{textcomp}%
\usepackage{manyfoot}%
\usepackage{booktabs}%
\usepackage{algorithm}%
\usepackage{algorithmicx}%
\usepackage{algpseudocode}%
\usepackage{listings}%

\usepackage{fancyhdr}

\fancypagestyle{firstpage}{
  \fancyhf{} 
  
  \fancyfoot[L]{\small{This preprint has not undergone peer review or any post-submission improvements or corrections. The Version of Record of this article is published in `KI - Künstliche Intelligenz`, and is available online at \url{https://doi.org/10.1007/s13218-026-00901-7}}}
  \fancyfoot[C]{\thepage}
}

\usepackage[colorinlistoftodos,prependcaption]{todonotes}

\usepackage{xspace}
\newcommand{\eg}{e.\,g.,\xspace}

\newcommand{\expris}{ExPrIS\xspace}

\usepackage{paralist}

\usepackage[acronym]{glossaries}
\glsdisablehyper   
\newacronym{3dssg}{3DSSG}{3D Semantic Scene Graph}
\newacronym{clip}{CLIP}{Contrastive Language-Image Pretraining}
\newacronym{gnn}{GNN}{Graph Neural Network}
\newacronym{xai}{XAI}{Explainable AI}

\usetikzlibrary{shapes.geometric, arrows.meta, positioning, fit, backgrounds, calc, shapes.misc}
\definecolor{exbg}{RGB}{235, 248, 235}
\definecolor{exnode}{RGB}{180, 235, 180}
\definecolor{mapblue}{RGB}{160, 215, 255}
\definecolor{liebg}{RGB}{253, 250, 235}
\definecolor{lienode}{RGB}{255, 245, 200}

\begin{document}

\title[ExPrIS project report]{ExPrIS: Knowledge-Level Expectations as Priors for Object Interpretation from Sensor Data}


\author*[1,2]{\fnm{Marian} \sur{Renz}}\email{marian.renz@dfki.de}

\author[1]{\fnm{Martin} \sur{Günther}}\email{martin.guenther@dfki.de}

\author[1]{\fnm{Felix} \sur{Igelbrink}}\email{felix.igelbrink@dfki.de}

\author[1]{\fnm{Oscar} \sur{Lima}}\email{oscar.lima@dfki.de}

\author[2,1]{\fnm{Martin} \sur{Atzmueller}}\email{martin.atzmueller@uos.de}


\affil[1]{\orgname{German Research Center for Artificial Intelligence}, \orgdiv{Research Department Cooperative and Autonomous Systems}, \orgaddress{\city{Osnabrück}, \country{Germany}}}
\affil[2]{\orgname{Osnabrück University, Institute of Computer Science}, \orgdiv{Semantic Information Systems Group}, \orgaddress{\city{Osnabrück}, \country{Germany}}}


\abstract{While deep learning has significantly advanced robotic object recognition, purely data-driven approaches often lack semantic consistency and fail to leverage valuable, pre-existing knowledge about the environment. This report presents the ExPrIS project, which addresses this challenge by investigating how knowledge-level expectations can serve as priors to improve object interpretation from sensor data. Our approach is based on the incremental construction of a \acrfull{3dssg}. We integrate expectations from two sources: contextual priors from past observations and semantic knowledge from external graphs like ConceptNet. These are embedded into a heterogeneous \acrfull{gnn} to create an expectation-biased inference process. This method moves beyond static, frame-by-frame analysis to enhance the robustness and consistency of scene understanding over time. The report details this architecture, its evaluation, and outlines its planned integration on a mobile robotic platform.}


\thispagestyle{firstpage} 



\maketitle


\section{Introduction}

Object recognition and semantic classification in sensor data like RGB or RGB-D have seen considerably advances in recent years---specifically due to significant methodological developments and improvements in machine-learning-based approaches. In particular, this is related to deep learning methods, which rely on purely data-driven approaches.

However, while such advanced object recognition approaches handle rich online sensor data well, they fail to exploit another important source of information: knowledge-level context about the environment. In robotics, for instance, such information is typically available to any autonomous system, especially those with a plan-based control layer. Specifically, this includes knowledge about environmental areas that are currently occluded or have not been recently observed.

In parallel, as a way to structure and reason about such environmental knowledge, \glspl{3dssg} \citep{Armeni2019-ip} have emerged in the past years as a symbolic abstraction of complex environments. Originally from the field of 2D image understanding \citep{Johnson2015-mw}, their application to 3D data for scene graph prediction \citep{Wald2020-yj} and semantic mapping \citep{Hughes2022-fg,Rosinol2020-wi} has led to significant advances in the field of robotics. More recently, the combination of \glspl{3dssg} with open set prediction models such as \gls{clip} \citep{Radford2021-mj} has produced even more powerful representations \citep{Gu2024-bq,Maggio2024-ca}.

However, the popular approaches to construct such an open set scene graph \citep{Gu2024-bq,Maggio2024-ca} still only rely on the power of the underlying deep neural networks, without considering prior observations from a sensor data stream or the knowledge level.

Hence, a symbolic representation of persistent objects, such as \glspl{3dssg}, together with expectations about where to be able to perceive them and powerful object detection functionality in sensor data should enable an integrated approach combining knowledge-driven as well as data-driven methods. This is the approach that we sketch in this paper -- connected to the \expris project.\footnote{\url{https://www.dfki.de/en/web/research/projects-and-publications/project/expris}} Here, the main research question is: \emph{What would be a fitting approach to processing expectations, derived from a knowledge level, as priors in modern scene graph generation methods?}

\section{Semantic Mapping with 3D Semantic Scene Graphs}
\label{sec:semantic-mapping-with-3dssgs}

As a foundational contribution of this project, we conducted a comprehensive survey on the state of the art in online knowledge integration for 3D semantic mapping \citep{Igelbrink2024-gg}.
The survey identifies a clear trend towards structured, symbolic environment representations, with \glspl{3dssg} emerging as a particularly powerful and flexible paradigm.
Guided by this analysis, the ExPrIS project adopts the \gls{3dssg} as its core data structure to systematically investigate the integration of knowledge-level expectations.

\begin{figure*}
    \centering
    \includegraphics[width=1.0\linewidth]{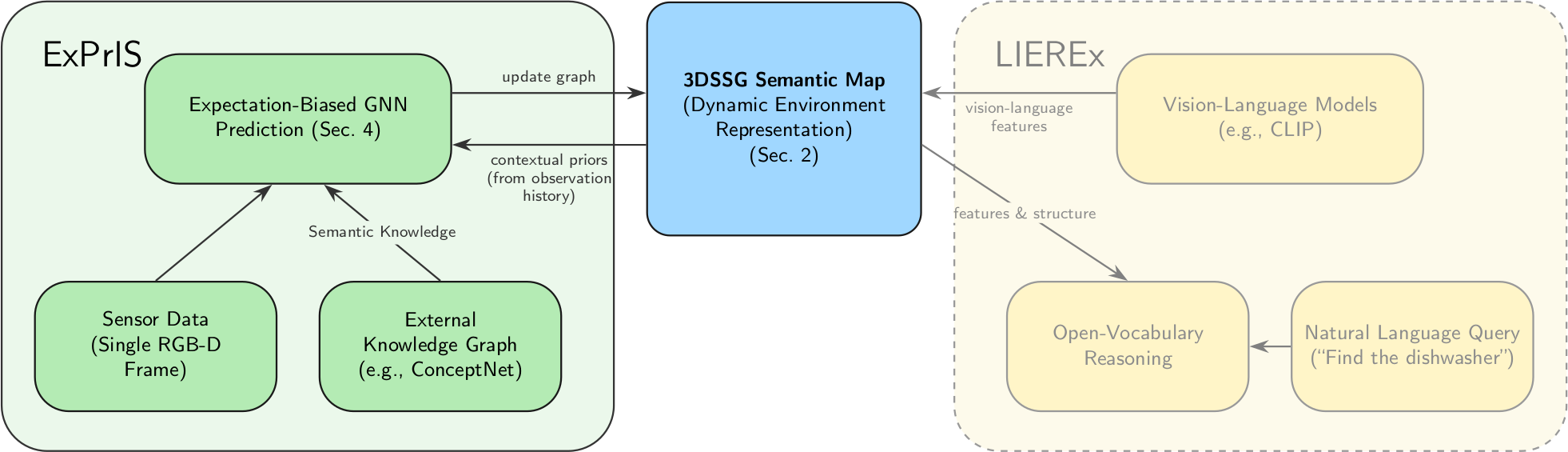}
    \caption{The ExPrIS/LIEREx semantic mapping architecture. This paper focuses on the ExPrIS and semantic map parts.}
    \label{fig:framework}
\end{figure*}

The ExPrIS project is complemented by its sister project, LIEREx (Language-Image Embeddings for Robotic Exploration); see Fig.~\ref{fig:framework}. While ExPrIS focuses on integrating structured knowledge to generate expectations, LIEREx extends this approach by incorporating modern vision-language models. This enables open-vocabulary queries and a more flexible, language-driven interaction with the environment. Both projects share the \gls{3dssg} as their foundational representation and are designed to be interoperable, combining symbolic knowledge (ExPrIS) with multimodal open-set perception (LIEREx).

\begin{figure}
    \centering
    \includegraphics[width=0.7\linewidth]{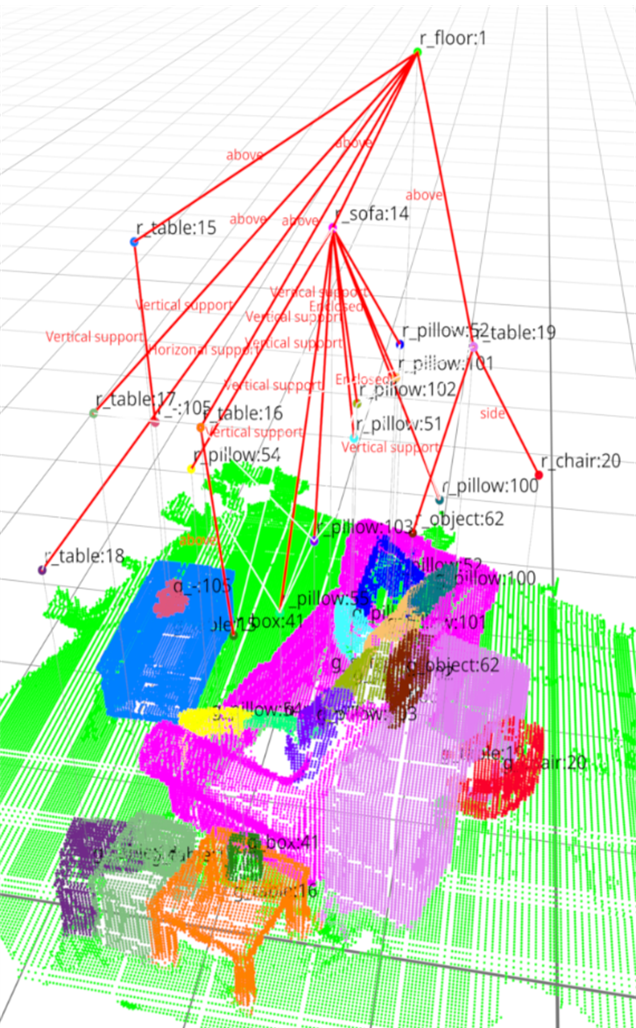}
    \caption{Hierarchical scene graph based on contact and support relationships.}
    \label{fig:hier-sg}
\end{figure}

We have based our semantic mapping framework on a dynamic \gls{3dssg}, which serves as the foundational environment representation for both ExPrIS and LIEREx. The general architecture of the \gls{3dssg} is inspired by the work of \citet{Rosinol2020-wi} as well as the Hydra framework \citep{Hughes2022-fg}.
The core component of the \gls{3dssg} is a heterogeneous graph organized in a hierarchy of multiple layers with different edge and node types. Each layer represents a different type of semantic concept ranging from low-level concepts (\eg individual objects) to higher-level concepts (\eg rooms and buildings). Besides per-layer edges to represent relationships between concepts of the same abstraction level, the \gls{3dssg} also allows for edges between nodes in different layers to represent meta-relations, \eg memberships. In Hydra and other preexisting works, the structure of layers is fixed to a rigid hierarchy of objects, places, rooms and buildings/floors which is suitable for mapping primarily simple organized indoor environments while allowing for scaling to larger environments consisting of many rooms in potentially multiple buildings as well \cite{Strader2024-cv}. However, this rigid structure limits adaptability to new domains and prevents the integration of dynamically inferred higher-order semantic concepts required to fully utilize existing domain-specific knowledge, which usually possesses a different underlying ontology as well as fully utilizing additional implicit knowledge sources such as \gls{clip} features. Another method is creating the underlying graph structure based on distance heuristics, \eg objects within a 0.5\,m radius of each other's bounding box center are connected by an edge in the scene graph \cite{Wald2020-yj}. This generalizes well and is easy to implement, but can create too densely connected subgraphs for scenes with many small objects in proximity to each other or fail to connect larger objects, if their size exceeds the distance threshold (0.5\,m in this example).
For our approach, we introduce a more modular and extensible layer design that separates spatial from semantic relations and supports dynamic inference of composite concepts. This flexibility allows the graph to evolve beyond static categories and incorporate domain-specific or task-driven abstractions. Our \gls{3dssg} is designed to be constructed incrementally from sensor data, with layers being updated in a bottom-up fashion. The scene graph structure is derived by detecting physical contact between objects and building a hierarchical graph structure based on support relationships (see Fig.~\ref{fig:hier-sg}).
Instead of utilizing truncated-signed-distance-function (TSDF)-based meshes to represent geometry, we currently employ a combination of bounding volumes and sparse voxel grids to represent low-level object instances and higher-level entities in the \gls{3dssg}. While TSDF grids provide high-quality surface reconstruction of geometry from 3D sensor data, they incur significant computational and memory overhead, limiting their scalability. Additionally, TSDF structures do not integrate well with existing deep learning frameworks without requiring additional modules and workarounds. Since our focus is on semantic reasoning rather than dense reconstruction, our choice allows for efficient storage and manipulation of 3D shapes and additional fast GPU-accelerated processing through integration into PyTorch while maintaining sufficient fidelity to support downstream tasks such as object recognition, spatial inference, and view planning.
The \gls{3dssg} implementation is designed to be modular and extensible, supporting future enhancements such as temporal reasoning, dynamic object tracking, and probabilistic inference as well as integration with other established frameworks such as the Hydra \gls{3dssg} mentioned earlier.
In future work, the currently separate heterogeneous \gls{3dssg} generation model described in Sec.~\ref{sec:expectation-biased-ml} will be fully integrated into this architecture, managing one or multiple layers of the \gls{3dssg} and enabling it to benefit from higher-order concepts and the knowledge encoded within them. Furthermore, in the LIEREx project, we plan on extending the hierarchical structure by introducing meta-nodes, which group semantically related objects. These meta-nodes can then be used as \textit{expectation} (see Sec.~\ref{sec:expectation-generation}) to assist scene graph prediction.

\section{Knowledge-Based Expectation Generation}
\label{sec:expectation-generation}

While mapping- or SLAM-based \glspl{3dssg} can be used as a structured scene representation, they lack semantic relationships between objects. Such relationships (\eg \emph{attached to}, \emph{standing on}, \emph{under}) can be modeled as edges between objects and deliver important semantic information for high-level planning. For this, we introduce an approach to incremental scene graph prediction, which utilizes \textit{expectations} during both training and inference.

We define an expectation to be any kind of additional information that is available prior to an inference step and can influence the outcome of the respective inference. For this work, we differentiate between two kinds of expectations:

\begin{enumerate}
    \item Context: information about the environment in which the prediction is taking place, such as results from prior predictions, \eg the scene graph constructed by predictions from previous frames
    \item Knowledge: static, general information about the domain, \eg formulated as a knowledge graph
\end{enumerate}

\begin{figure*}
    \centering
    \includegraphics[width=0.8\textwidth]{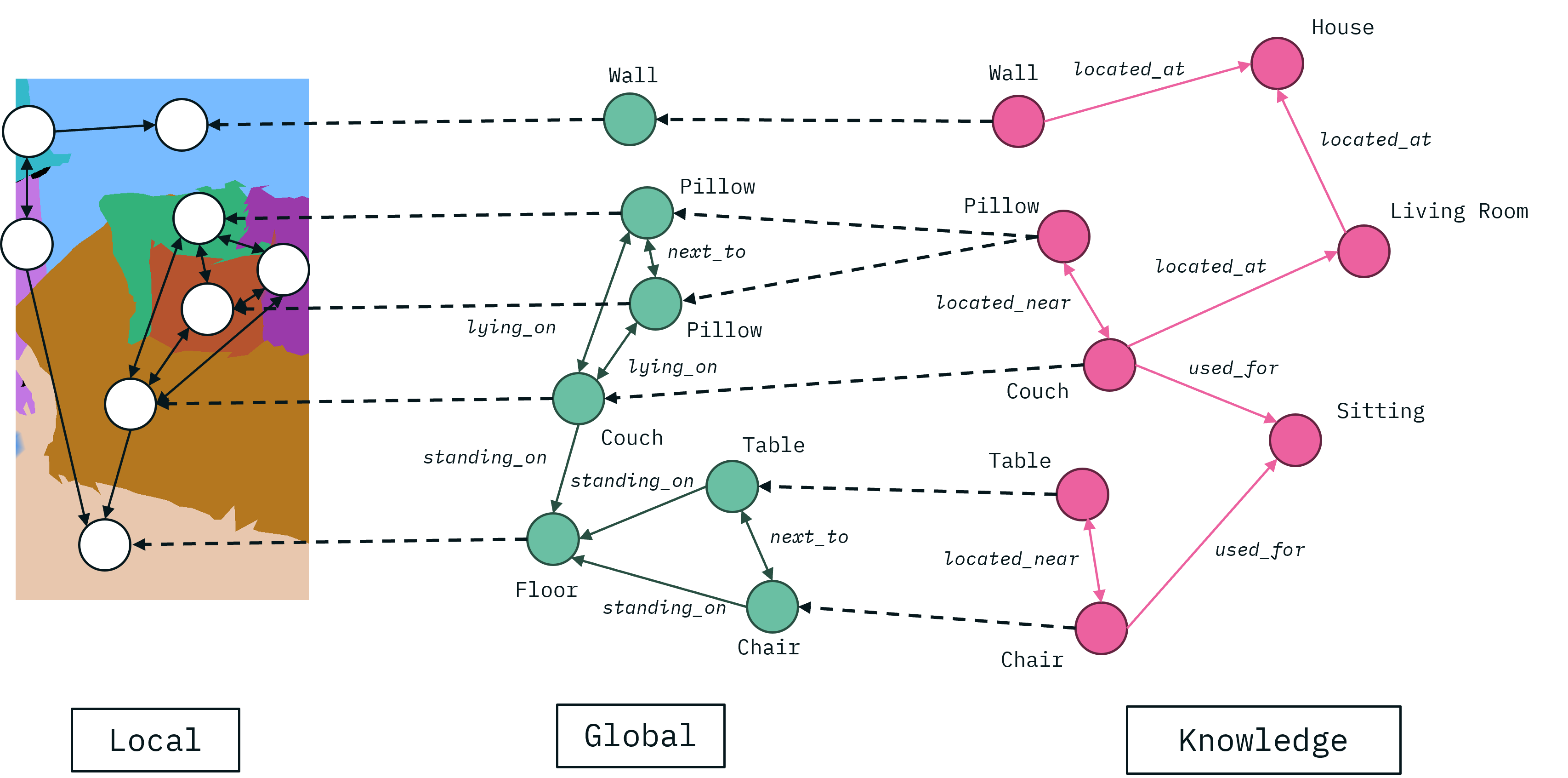}
    \caption{\gls{3dssg} model for expectation integration}
    \label{fig:hetero-graph}
\end{figure*}

Fig.~\ref{fig:hetero-graph} shows the presented concept of expectation integration for \gls{3dssg} prediction from RGB-D sensor data. The goal is to incrementally predict a global \gls{3dssg} by accumulating local scene graphs predicted from single, segmented RGB-D frames. Nodes in this graph represent the segmented objects, and edges represent the semantic or spatial relationships between these objects. The presence of a potential edge is estimated using a simple distance threshold between object segments in the projected depth frame as heuristic (see Sec.~\ref{sec:semantic-mapping-with-3dssgs}). Node and edge classes are then predicted using a \gls{gnn}. In the current state, this global \gls{3dssg} is constructed separately for \gls{gnn} training and not by using the mapping framework described in Sec.~\ref{sec:semantic-mapping-with-3dssgs} but will be integrated as described in Sec.~\ref{sec:evaluation}.

The current state of this accumulated, global \gls{3dssg} serves as the first type of expectation to be integrated in each local graph prediction. It can be thought of as a context or prior observations, assisting the prediction. The integration occurs by matching segments from the local frame to already observed geometric instances and connecting matching nodes with an edge (see Fig.~\ref{fig:hier-sg}). This way, information from the global graph can traverse into the local graph by the \gls{gnn}’s message passing process.

Using the same principle, the global nodes can be connected to a static knowledge graph using the predicted node classes. This knowledge layer can be interpreted as external prior knowledge about the environment, the second type of expectations.
To this end, we chose the common-sense knowledge graph ConceptNet \citep{Speer2017-os} as a source for prior knowledge. Common indoor 3D datasets usually consist of everyday objects and scenes, making common-sense knowledge graphs a suitable source for additional knowledge. However, ConceptNet contains a lot of noise by \eg numerous nodes and relations from linguistic contexts (\eg metaphors or abstract concepts) which are not useful for physical scene understanding and \gls{3dssg} generation. This and the sheer size of ConceptNet requires the extraction of subgraphs based on the desired nodes and relationships.

The pipeline for extracting subgraphs from ConceptNet consists of a breadth-first search based on a set of node classes and specified relationship types for n hops. Furthermore, various graph embedding algorithms have been implemented to generate additional feature vectors for the knowledge graph required for the message passing process, providing alternatives to the pre-computed Numberbatch \citep{Speer2017-os} graph embeddings already provided by ConceptNet.

\section{Expectation-Biased Machine Learning}
\label{sec:expectation-biased-ml}

Building on the concept of expectations introduced in Sec.~\ref{sec:expectation-generation}, we embed prior observations and semantic knowledge directly into the learning process for incremental 3D scene graph prediction. The objective is to move beyond static, frame-by-frame perception and enable context-aware, expectation-biased inference that improves robustness and semantic consistency over time.

For incremental prediction, the model processes individual RGB-D frames sequentially, constructing partial local graphs from individual observations in the frame and integrating them into a growing global graph. This setup reflects real-world robotic perception tasks, where complete scene reconstructions are rarely available upfront.
While Sec.~\ref{sec:expectation-generation} introduced the principle of linking local observations to a global context, this section focuses on how this linkage is exploited during learning. Instead of treating the global graph as a passive memory, we integrate it as an active component in the message-passing process of a \gls{gnn}. The global layer aggregates historical observations and semantic priors, while the local layer represents the current frame. Cross-layer edges allow for information flow between the layers, enabling the model to refine predictions for partially observed objects and relationships by leveraging accumulated context. This design transforms the global graph from a static expectation source into a dynamic biasing mechanism during inference.

Global nodes are enriched with prior-based features, such as previously predicted class labels or \gls{clip} embeddings \citep{Radford2021-mj}. Unlike earlier approaches that encode global context as fixed vectors, our method integrates these priors directly into the heterogeneous graph structure, ensuring that expectations adaptively influence both node and edge classification.
The architecture employs a heterogeneous \gls{gnn} (GraphSAGE \citep{Hamilton2017-nc} or HGT \citep{Hu2020-gs}) capable of handling multiple node and edge types. Node features combine PointNet-based \citep{Qi2017-wl} geometric feature vectors with additional handcrafted descriptors, while edge features capture relative spatial configurations. The training objective is a composite loss over node and edge predictions in the local graph, scaled to account for class imbalance. Importantly, only local predictions contribute to the loss, ensuring that the global graph serves as a bias rather than a direct supervision signal. A more detailed overview of the presented architecture and evaluation can be found in our paper \cite{Renz2025-uk}; code is available on GitHub.\footnote{\url{https://github.com/m4renz/incremental-scene-graph-prediction}}

\section{Planned Integration and Evaluation on a Robot Platform}
\label{sec:evaluation}

To bridge the gap between offline dataset analysis and real-world application, the final stage of the ExPrIS project is dedicated to integrating and evaluating the full pipeline on an embodied agent. This step is essential in order to validate the robustness and practical benefits of our expectation-biased framework in a dynamic environment subject to real-world sensor noise and unmodeled complexities.

\begin{figure}
    \centering
    \includegraphics[width=0.7\linewidth]{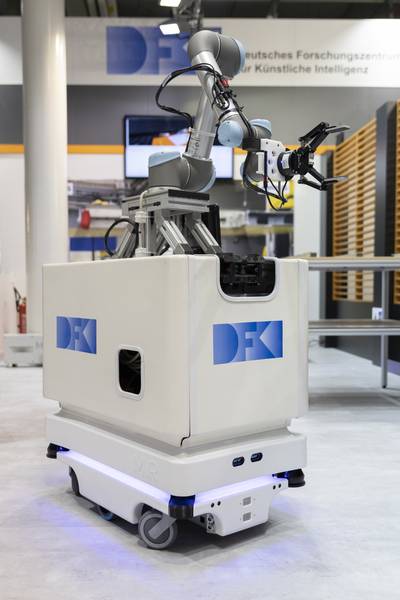}
    \caption{The Mobipick robot}
    \label{fig:mobipick}
\end{figure}

The planned experiments will be conducted on the Mobipick \citep{Lima2023icaps} mobile manipulator (Fig.~\ref{fig:mobipick}). This platform combines a MiR100 mobile base with a UR5 robotic arm and a Robotiq gripper, providing capabilities for both navigation and object manipulation. Its end-effector-mounted RGB-D camera will serve as the primary sensor, delivering the rich data streams required for our perception pipeline.

The Mobipick platform will be equipped with the complete software architecture presented in this paper. The system's primary objective will be to incrementally build a \gls{3dssg} (Sec.~\ref{sec:semantic-mapping-with-3dssgs}) of its environment in real time. The core of the evaluation will focus on the performance of the expectation-biased \gls{gnn} model (Sec.~\ref{sec:expectation-biased-ml}), which will process live sensor data, leveraging both contextual and semantic priors (Sec.~\ref{sec:expectation-generation}) to continuously refine its scene interpretation.

To test the system's capabilities, we will employ a ``tidy-up'' scenario. In this task, the robot will first explore a lab environment to build a semantic map. Subsequently, it will be tasked with identifying and relocating objects that are ``out of place'', requiring it to reason about both the current and the correct semantic context of objects. This scenario is designed to rigorously test the core value of ExPrIS: improving semantic consistency and robustness in perception over longer-term robot operation.

\section{Discussion and Outlook}

Through the real-world experiments proposed in the previous section, we hypothesize that the expectation-biased approach will demonstrate clear advantages over context-free, frame-by-frame perception models. We expect to observe a significant improvement in semantic consistency, where the system's understanding of objects remains stable and coherent over time, even in the face of partial observations or perceptual ambiguities.
The ``tidy-up'' scenario is specifically designed to test whether the robot can resolve such ambiguities by leveraging its memory of past observations---a key feature enabled by the global graph structure.

Looking beyond this immediate validation, future work will extend the incremental scene graph prediction architecture in several key directions. We plan to more deeply incorporate external knowledge graphs (as introduced in Sec.~\ref{sec:expectation-generation}), moving beyond their use as simple priors towards enabling more complex, symbolic knowledge processing. This will be critical for our long-term goals of large-scale semantic mapping and \gls{xai}. In parallel to the core prediction task, we have already begun investigating methods for graph \gls{xai} to assess the effect and extent of the knowledge graph's influence on local scene graph predictions. A comprehensive evaluation of the embodied system, including the full integration of the open-vocabulary query capabilities from the sister project LIEREx, is currently underway, and the results will be the subject of a forthcoming publication.


\backmatter


\bmhead{Acknowledgements}

This work is supported by the ExPrIS project through a grant from the German Federal Ministry of Research, Technology and Space (BMFTR) with Grant Number 16IW23001.
The DFKI Niedersachsen (DFKI NI) is sponsored by the Ministry of Science and Culture of Lower Saxony and the Volkswagen Stiftung.

\section*{Declarations}


The authors have no competing interests to declare that are relevant to the content of this article.

\bibliography{literature_expris_ki}

\end{document}